\documentclass{article}
\usepackage{spconf,amsmath,graphicx}

\usepackage{multirow}
\usepackage{makecell}
\usepackage{hyperref}

\usepackage{subcaption}

\title{Automatically Identifying Language Family from Acoustic Examples in Low Resource Scenarios}

\name{Peter Wu, Yifan Zhong, Alan W Black}
\address{Language Technologies Institute, Carnegie Mellon University, PA, USA}
\begin{document}
%
\maketitle
\begin{abstract}
Existing multilingual speech NLP works focus on a relatively small subset of languages, and thus current linguistic understanding of languages predominantly stems from classical approaches. In this work, we propose a method to analyze language similarity using deep learning. Namely, we train a model on the Wilderness dataset and investigate how its latent space compares with classical language family findings. Our approach provides a new direction for 
cross-lingual data augmentation in any speech-based NLP task.\footnote{\href{https://github.com/peter-yh-wu/multilingual}{https://github.com/peter-yh-wu/multilingual}}
\end{abstract}
%


\section{Introduction}

As NLP systems become stronger, it is important to make them available to everyone around the world. While there are thousands of languages in the world, strong NLP systems are currently only available in a small subset of these languages. Since state-of-the-art NLP systems typically have a data requirement, it is difficult if not infeasible to currently build them for low resource languages. As more and more language data is being collected, one potential solution for building low-resource NLP systems is leveraging data from related languages. Thus, a deeper understanding of the relationship between languages can help identify new data augmentation solutions for multilingual NLP systems. Direct applications include cross-lingual training and adaptation for tasks like zero-shot text-to-speech (TTS).

Existing works on language similarity classify languages using genetic information and typological deductions by expert linguists \cite{ethnologue, glottolog}. Standards like the International Phonetic Alphabet (IPA) have also been developed to unify the representation of languages.
While there exist multilingual standards, more detailed language analysis typically varies between linguist. 
Algorithmic analysis from raw language data provides a potential approach towards universal language analysis.

In this paper, we propose a method to analyze language similarity using deep learning to extract language information directly from raw speech data. We proceed by discussing related works in Section \ref{sec:related_works}, our proposed approach in Section \ref{sec:pro}, experiments in Section \ref{sec:exp}, and conclude in Section \ref{sec:conc}.

\section{Related Works}
\label{sec:related_works}

\subsection{Language Classification}

Existing works classify languages into family trees based on a combination of genealogical and typological information decided by expert linguists \cite{ethnologue, glottolog}.
Since classification decisions vary between linguists, language family trees vary noticeably between sources, which we briefly discuss in section \ref{metrics}. In order to mitigate linguist bias, we propose a neural approach that extracts language information directly from raw speech data.





\subsection{Multilingual Speech Processing}

Audio data provide another way to analyze languages, whether from phonemes or raw waveforms. Phoneme-based approaches involve either defining a new set of acoustic words or using existing standards. While the former approach can yield  downstream performance competitive to the latter, existing work is primarily restricted to monolingual tasks \cite{lee2015unsupervised, kamper_phoneme}. Standards like IPA define a set of phonemes aimed to support all spoken languages, and tools like PHOIBLE provide data representing languages following these standards \cite{phoible}. In section \ref{metrics}, we propose ways to measure language similarity using these tools.

Languages can also be analyzed directly from raw speech data. There currently exists numerous statistical and deep learning techniques to process audio from multiple languages \cite{multilingual_asr_google, multilingual_asr_edinburgh, multilingual_asr_cnnlstm, heigold2013multilingual, black2019cmu_wilderness}.
A few of these works analyze the relationship between languages, but only on a small number of languages
\cite{conneau2020xlsr, pigoli2018statistical}. In this work, we propose an approach that can compare hundreds of languages and generalize to many more.


\subsection{Massively Multilingual NLP}

Increased data and compute availability has enabled the NLP community to process much more than just a handful of languages \cite{xtreme, adams2019massively_shinji_asr, ammar2016massively}. While most publicly available multilingual data is either text or high-resource speech, datasets like Wilderness and Common Voice provide opportunities to analyze multilingual speech data in numerous languages \cite{black2019cmu_wilderness, ardila2019common}. In this work, we train a neural model on speech data from hundreds of languages in Wilderness, and use the resulting model to analyze the relationship between languages. Additionally, we test the downstream potential of our approach through a zero-shot TTS task.

\section{Proposed Approach}
\label{sec:pro}

\subsection{Dataset}

\begin{figure}[h]
\centering
\caption{$195$-language Wilderness subset color-coded by language family}
\includegraphics[width=8cm]{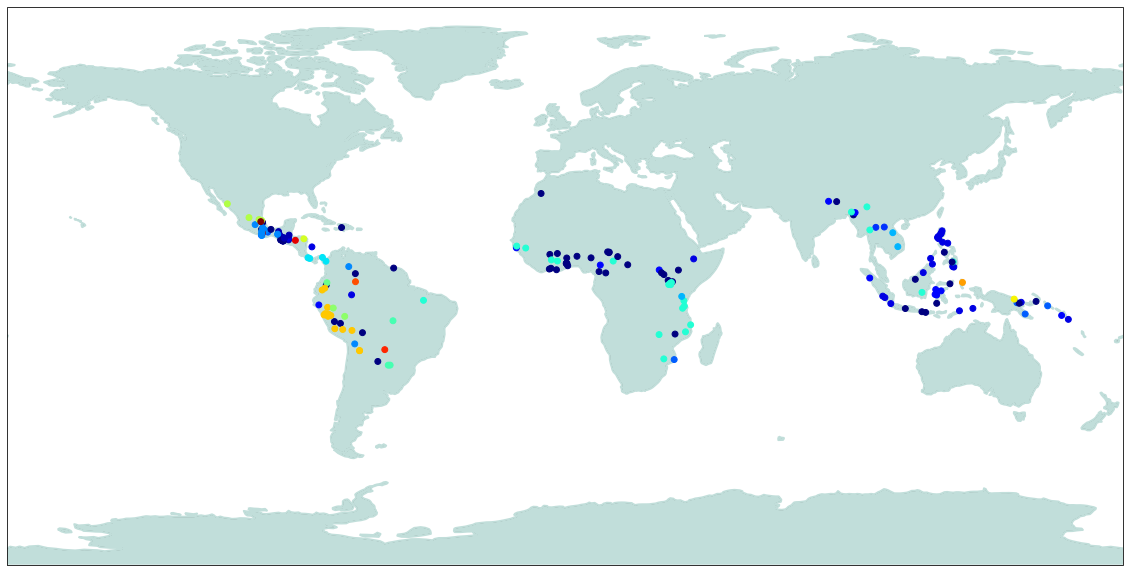}
\label{fig:lang_fams}
\end{figure}

All our experiments are conducted on the Wilderness dataset. Specifically, we use the languages with mel-cepstral distortion (MCD) less than $5.5$. Since these languages predominantly lie in the Americas, Africa, and Southeast Asia, this work focuses on languages from these regions. This yields $195$ languages, with $22.8\pm 5.9$ hours of speech per language. Figure \ref{fig:lang_fams} visualizes the location of these languages.

\subsection{Model}

Our model feeds 80-bin mel-scale spectrogram features into an LSTM with hidden dimension $256$, performs a max pooling operation on the frame dimension, and then feeds the resulting $256$-dimensional vector into a fully connected layer with output dimension 64, then a ReLU activation, and finally another fully connected layer with output dimension 195. 
During training, we use $90\%$ of the data and an Adam optimizer with learning rate $10^{-3}$.
We use this model to output a 64-dimensional vector embedding for each language. Namely, we take the outputs of the first fully connected layer for every data point in the test set and compute averages for each language. Unless mentioned otherwise, we generate embeddings with a model that has a test accuracy of $0.7049$.

\subsection{Metrics}
\label{metrics}

\subsubsection{Geographical Distance}
\label{sec:metric:geo}

We compare the distances between language embeddings with the distances between the geographical locations of the corresponding languages. Specifically, for each language, we compute the Euclidean distance of its language embeddings with those of the 194 other languages and then calculate the Pearson correlation coefficient between the embedding distances and the corresponding geographical distances. We then take the average and standard deviation of these 195 correlation values to obtain our geographical distance correlation metric. We write this metric $m$ mathematically as 
\begin{equation} \label{eq:global_cor}
m = \mu_c \pm \sigma_c, c = \frac{\text{cov}(e_{i,:}, g_{i,:})}{\sigma_{e_{i,:}} \sigma_{g_{i,:}}},
\end{equation}
where $e_{i,j}$ is the Euclidean distance between languages $i$ and $j$, $g_{i,j}$ is the corresponding geographical distance, cov is the covariance function, $\mu_c$ is the sample mean of $c$, and $\sigma_c$ is the sample standard deviation of $c$.

Additionally, in order to assess how our embedding approach performs locally, we compare each language with only those within a radius of $r$ kilometers, where $r = 500$ unless mentioned otherwise. Mathematically, this is equivalent to replacing ``$:$'' in Equation \ref{eq:global_cor} with ``$\{j | d(i,j) < r\}$'', where $d(i,j)$ is the geographical distance between languages $i$ and $j$.
We refer to Equation \ref{eq:global_cor} as global correlation (no pun intended), and the nearby version as local correlation.

\subsubsection{Language family tree}
\label{sec:pro:tree}

\begin{figure}[h]
\centering
\caption{Top two levels of Cariban language family tree based on three different data sources.}
     \begin{subfigure}[b]{0.32\textwidth}
         \centering
         \includegraphics[width=\textwidth]{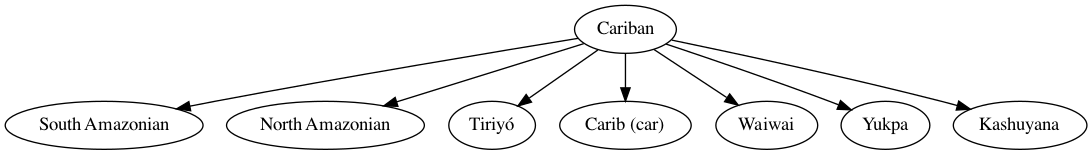}
         \caption{Ethnologue}
         \label{fig:ethnologue}
     \end{subfigure}
     \hfill
     \begin{subfigure}[b]{0.13\textwidth}
         \centering
         \includegraphics[width=\textwidth]{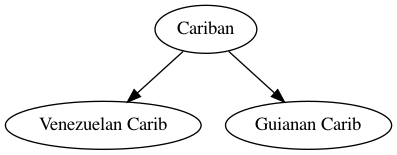}
         \caption{Wikipedia}
         \label{fig:wikipedia}
     \end{subfigure}
     \hfill
     \begin{subfigure}[b]{0.45\textwidth}
         \centering
         \includegraphics[width=\textwidth]{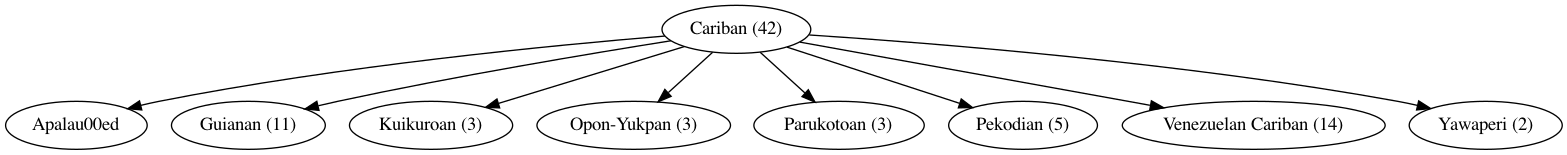}
         \caption{Glottolog}
         \label{fig:glottolog}
     \end{subfigure}
\label{fig:cariban_trees}
\end{figure}

We compare three different types of language family trees in order to devise our tree-based metric. We build our first set of trees using Ethnologue \cite{ethnologue}. Specifically, we convert the Classification label of each language into a branch with the root node being the language family, other non-leaf nodes being language groups, and the leaf being the language. We use a similar approach on the respective language family data in Wikipedia and Glottolog to build two more sets of trees \cite{glottolog}. Figure \ref{fig:cariban_trees} visualizes subtrees from each of the three approaches.

Table \ref{table:tree_stats} summarizes a few notable language categorization statistics for the three approaches. We refer to a language as single if it is the only one amongst the 195 languages that is in its family. Since the Ethnologue-based trees have the least number of single languages, we choose to use that approach as the default one for our tree-based metrics.

\begin{table}[h]
\fontsize{9}{11}\selectfont
\centering
\caption{Language family tree statistics. Leaves that do not correspond to one of the 195 languages are removed. We observe that among the resulting three forests, Ethnologue has the least number of family trees and single languages.}
\setlength\tabcolsep{2.0pt}
\begin{tabular}{ l || c c c }
\Xhline{1\arrayrulewidth}
{\sc Tree}  & {Non-Single} & {Families} & {Isolates} \\
\Xhline{0.5\arrayrulewidth}
Ethnologue & $185 (94.87\%)$ & $24$ & $1$ \\
Wikipedia & $177 (90.77\%)$ & $37$ & $2$ \\
Glottolog & $181 (92.82\%)$ & $34$ & $3$ \\
\Xhline{1\arrayrulewidth}
\end{tabular}
\label{table:tree_stats}
\vspace{-0mm}
\end{table}

We evaluate our embedding approach by assessing its ability to match language similarity measures defined by the trees. Specifically, for a set of trees and a threshold value $k$, we compute the percentage of languages that have a family member among their $k$ closest neighbors in the embedding space, where family member refers to a language in the same tree and distance is measured using the Euclidean norm. We report results using this metric in Section \ref{sec:exp:tree}.



\subsection{Outlier Analysis}
\label{sec:pro:outlier}

We explore using our embedding approach to identify outlier languages. Namely, for each language, we calculate its average embedding distance to other languages in the same family. We consider the languages with highest average embedding distances as outliers, as detailed in Section \ref{sec:exp:outlier}. We use the Ethnologue tree to define language families based on the findings in Section \ref{sec:pro:tree}.

\subsection{TTS Analysis}
\label{sec:pro:tts}

To analyze how our embedding approach performs in downstream tasks, we conduct a zero-shot TTS task. Specifically, for each Wilderness language, we train a TTS model only using data from a similar but different language, and evaluate the model performance on data from the original language. We choose the similar language using three approaches: 1. embedding distance, 2. geographical distance, and 3. phoneme set. In the third approach, we use the embedding distance between binary bag-of-phonemes vectors generated using PHOIBLE \cite{phoible}. We use the Random Forest model for TTS as described in Wilderness \cite{black2019cmu_wilderness, Black2015RandomFF}. Experimental details are described in Section \ref{sec:exp:tts}.

\section{Experiments}
\label{sec:exp}

\subsection{Geographical Distance}
\label{sec:exp:geo}

Table \ref{table:cor} describes the global and local correlations between geographical distance and embedding distance for models with difference language classification accuracies. We observe that exposure to more speech data indeed helps the model learn more accurate distances between languages. Additionally, our best model achieves 0.6533 correlation with geographical distance when comparing languages within 500 kilometers from each other. The lower global correlation suggests that geographical distance may not be as indicative of language similarity as languages are further apart.


\begin{table}[h]
\fontsize{9}{11}\selectfont
\centering
\caption{Correlations between embedding distance and geographical distance for models with different language classification accuracies.
Our best model achieves 0.65 correlation with geographical distance when comparing languages within 500 kilometers from each other. Lower global correlation suggests that larger distances, e.g. due to bodies of water, may not necessarily mean that languages are more dissimilar.}
\setlength\tabcolsep{2.0pt}
\begin{tabular}{ l || c c }
\Xhline{1\arrayrulewidth}
{\sc Accuracy}  & {Global ($\mu \pm \sigma$)} & {Local ($\mu \pm \sigma$)} \\
\Xhline{0.5\arrayrulewidth}
0.0140 & $0.1915 \pm 0.2827$ & $0.5345 \pm 0.3603$ \\
0.1195 & $0.2376 \pm 0.2571$ & $0.5704 \pm 0.3516$ \\
0.3250 & $0.2768 \pm 0.2467$ & $0.6199 \pm 0.3233$ \\
0.5296 & $0.2907 \pm 0.2320$ & $0.6315 \pm 0.3112$ \\
0.7049 & $\mathbf{0.2982 \pm 0.2196}$ & $\mathbf{0.6533 \pm 0.2948}$ \\
\Xhline{1\arrayrulewidth}
\end{tabular}
\label{table:cor}
\vspace{-0mm}
\end{table}



\subsection{Language Family Tree}
\label{sec:exp:tree}

Table \ref{table:tree_k} summarizes our results using the tree-based metric described in Section \ref{sec:pro:tree}. We omitted languages that have no family members among the other 194 languages since their average embedding distance to other family members is undefined, yielding 185, 177, and 181 languages for the Ethnologue-, Wikipedia-, and Glottolog-based approaches, respectively. For all three approaches, we observe that for over $90\%$ of the languages, our embedding approach matches languages with at least one family member among the $k=16$ closest neighbors in the embedding space, and does so for nearly all of the languages when $k=64$.

\begin{table}[h]
\fontsize{9}{11}\selectfont
\centering
\caption{Performance of our embedding approach based on our Section \ref{sec:pro:tree} tree-based metric for three different tree types, reported as percentages. We observe that for over $90\%$ of the approximately $180$ languages, our embedding approach matches languages with at least one family member among the $16$ closest neighbors in the embedding space.}
\setlength\tabcolsep{2.0pt}
\begin{tabular}{ l || c c c c c c }
\Xhline{1\arrayrulewidth}
{\sc Tree Type}  & {$k=2$} & {$k=4$} & {$k=8$} & {$k=16$} & {$k=32$} & {$k=64$} \\
\Xhline{0.5\arrayrulewidth}
Ethnologue & $58.38$ & $69.73$ & $78.92$ & $88.11$ & $95.14$ & $97.30$ \\
Wikipedia & $71.19$ & $78.53$ & $84.18$ & $89.27$ & $94.92$ & $97.74$ \\
Glottolog & $56.35$ & $63.54$ & $70.17$ & $79.01$ & $88.40$ & $95.03$ \\
\Xhline{1\arrayrulewidth}
\end{tabular}
\label{table:tree_k}
\vspace{-0mm}
\end{table}

\subsection{Outlier Analysis}
\label{sec:exp:outlier}

Table \ref{table:outlier} describes the top five outliers identified in our 195-language Wilderness subset using our embedding approach, detailed in Section \ref{sec:pro:outlier}. 
We omitted languages that have no family members among the other 194 languages since their average embedding distance to other family members is undefined, yielding 185 languages.
The furthest outlier, Tilantongo Mixtec (XTDTBL), is also identfied linguistically as noticeably different from other languages in its family \cite{mixtec_outlier}.
Additionally, two of the other top five outliers have no nearby languages in their family, and thus separation due to geographical distance may have caused these languages to be phonologically far from the rest of their families.
For the other two top five outliers, the majority of nearby languages are not in their families, suggesting that both of these may exhibit qualities differing from other languages in their family, as is the case with Tilantongo Mixtec.

\begin{table}[h]
\fontsize{9}{11}\selectfont
\centering
\caption{Outliers, defined as the languages with greatest embedding distance to other languages in their family. We report the number of other languages within 500 kilometers, within the same family, and both, as well as the average embedding distance to family members. The first row provides the mean and standard deviation across languages. Existing linguistic work has confirmed that the furthest outlier is not close other languages in its family \cite{mixtec_outlier}.
}
\setlength\tabcolsep{2.0pt}
\begin{tabular}{ l || c c c c }
\Xhline{1\arrayrulewidth}
{\sc Lang. ID}  & {Nearby Family} & {Family} & {Nearby} & {Distance} \\
\Xhline{0.5\arrayrulewidth}
$\mu \pm \sigma$ & $6.2 \pm 5.1$ & $26.7 \pm 20.9$ & $22.7 \pm 15.8$ & $11.6 \pm 1.9$ \\
\Xhline{0.5\arrayrulewidth}
XTDTBL & $10$ & $12$ & $47$ & $17.9$ \\
CWEPBT & $10$ & $22$ & $18$ & $17.9$ \\
APRWBT & $0$ & $25$ & $6$ & $16.9$ \\
MMSBSG & $9$ & $57$ & $50$ & $16.5$ \\
SHIRBD & $0$ & $57$ & $0$ & $16.3$ \\
\Xhline{1\arrayrulewidth}
\end{tabular}
\label{table:outlier}
\vspace{-0mm}
\end{table}

\subsection{TTS Analysis}
\label{sec:exp:tts}

To compare the zero-shot TTS performances using our embedding distance approach and the geographical distance approach described in Section \ref{sec:pro:tts}, we experiment on a subset of the 195 Wilderness languages. In order to avoid including isolate languages, we chose languages with the most Ethnologue family members.
This amounted to 36 languages spanning 4 different families. For each language, we retrieved its $k$th closest neighbor within our 195-language subset using both distance methods, where $k\in\{1,5,10,50\}$. For each neighbor, we train a Random Forest TTS model using all of the Wilderness data for that language \cite{black2019cmu_wilderness, Black2015RandomFF}. We also similarly train a TTS model for each of the 36 languages. Then, for each of the 36 languages, we synthesize the first 100 lines in Wilderness for that language using the respective TTS model as well as the neighbor TTS models. For each neighbor, we measure zero-shot TTS performance using the mel-cepstral distortion between its generated waveforms and those from the TTS model trained on the original language. Table \ref{table:tts} summarizes these results. We observe that our embedding approach generally outperforms the geographical distance approach. These results are not completely unexpected, as even though our embedding and TTS approaches optimize for different objectives, their core features are both spectral in nature. TTS performance drops faster with higher $k$ for the geo-distance approach, matching our global versus local hypothesis discussed in Section \ref{sec:exp:geo} and suggesting that our embedding approach is able to identify larger sets of similar languages than the geographical distance approach.


\begin{table}[h]
\fontsize{9}{11}\selectfont
\centering
\caption{Zero-shot TTS performance using embedding and geographical distances, where we train TTS models on the $k$th closest neighboring language. For each $k$, we report the mean MCD value averaged across the 36 language experiments. We observe that our embedding approach generally outperforms the geographical distance approach. TTS performance drops faster with higher $k$ for the latter approach, suggesting that our embedding approach is able to identify larger sets of similar languages than the geographical distance approach.}
\setlength\tabcolsep{2.0pt}
\begin{tabular}{ l || c c c c }
\Xhline{1\arrayrulewidth}
{\sc Approach}  & {$k=1$} & {$k=5$}  & {$k=10$} & {$k=50$} \\
\Xhline{0.5\arrayrulewidth}
Embedding & $5.12$ & $\mathbf{5.35}$ & $\mathbf{5.20}$ & $\mathbf{5.57}$ \\
Geo-Distance & $\mathbf{5.00}$ & $5.48$ & $5.73$ & $5.75$ \\
\Xhline{1\arrayrulewidth}
\end{tabular}
\label{table:tts}
\vspace{-0mm}
\end{table}

We also compare the zero-shot TTS performances using our embedding approach and the bag-of-phonemes approach described in Section \ref{sec:pro:tts}. Since PHOIBLE only contains data for 94 out of our 195 languages, we chose languages with the most family members among these 94, amounting to 39 languages spanning 6 different families. Table \ref{table:tts_ph} summarizes our results, which also suggest that our embedding approach is able to identify larger sets of similar languages than the bag-of-phonemes approach.

\begin{table}[h]
\fontsize{9}{11}\selectfont
\centering
\caption{Zero-shot TTS performance using embedding and bag-of-phonemes distances. We observe that our embedding approach generally outperforms the phoneme-based one.}
\setlength\tabcolsep{2.0pt}
\begin{tabular}{ l || c c c c }
\Xhline{1\arrayrulewidth}
{\sc Approach}  & {$k=1$} & {$k=5$}  & {$k=10$} & {$k=50$} \\
\Xhline{0.5\arrayrulewidth}
Embedding & $5.13$ & $\mathbf{5.33}$ & $\mathbf{5.60}$ & $\mathbf{5.98}$ \\
Phonemes & $\mathbf{5.01}$ & $5.77$ & $5.76$ & $6.03$ \\
\Xhline{1\arrayrulewidth}
\end{tabular}
\label{table:tts_ph}
\vspace{-0mm}
\end{table}

\section{Conclusion and Future Directions}
\label{sec:conc}

In this work, we investigate a new way to find the correlation between acoustics and language family. An immediate application of our approach is zero-shot TTS, which we show can yield performant synthesizers. In the future, we plan to extend our acoustic distance approach to methods that make existing language family trees more consistent at the phonological level.




\bibliographystyle{IEEEbib}

\bibliography{refs}


\end{document}